\pdfoutput=1
\documentclass[11pt]{article}

\usepackage{ACL2023}

\usepackage{times}
\usepackage{latexsym}

\usepackage[T1]{fontenc}
\usepackage[utf8]{inputenc}

\usepackage{microtype}

\usepackage{inconsolata}

\usepackage{xspace}
\usepackage{graphicx}
\usepackage{amssymb}
\usepackage{multirow}
\usepackage{multicol}
\usepackage{array}
\usepackage{booktabs}
\usepackage{algorithm}
\usepackage{algorithmic}
\usepackage{amsmath}
\usepackage{makecell}
\usepackage{amsfonts}
\usepackage{bbold}

\newcommand\atomic{\textsc{Atomic}\xspace}
\newcommand\cometTT{\textbb{COMET}$_{ours}$\xspace}
\newcommand\atomicTT{Dense-\textsc{Atomic}\xspace}
\newcommand\comet{\textbb{COMET}\xspace}

\title{\atomicTT{}: Towards Densely-connected~\atomic{} \\with High Knowledge Coverage and Massive Multi-hop Paths}

\author{Xiangqing Shen, Siwei Wu, and Rui Xia\thanks{*Corresponding author} \\
        School of Computer Science and Engineering, \\ Nanjing University of Science and Technology, China \\
        \texttt{\{xiangqing.shen, wusiwei, rxia\}@njust.edu.cn}}
        
\begin{document}

\maketitle

\begin{abstract}

\atomic{} is a large-scale commonsense knowledge graph (CSKG) containing everyday \emph{if-then} knowledge triplets, i.e., \{\emph{head event}, relation, \emph{tail event}\}.
The one-hop annotation manner made \atomic{} a set of independent bipartite graphs, which ignored the numerous links between events in different bipartite graphs and consequently caused shortages in knowledge coverage and multi-hop paths. 
In this work, we aim to construct \atomicTT{} with high knowledge coverage and massive multi-hop paths.
The events in \atomic{} are normalized to a consistent pattern at first.
We then propose a CSKG completion method called Rel-CSKGC to predict the relation given the \emph{head event} and the \emph{tail event} of a triplet, and train a CSKG completion model based on existing triplets in \atomic{}.
We finally utilize the model to complete the missing links in \atomic{} and accordingly construct \atomicTT{}.
Both automatic and human evaluation on an annotated subgraph of \atomic{} demonstrate the advantage of Rel-CSKGC over strong baselines.
We further conduct extensive evaluations on \atomicTT{} in terms of statistics, human evaluation, and simple downstream tasks, all proving \atomicTT{}'s advantages in Knowledge Coverage and Multi-hop Paths.
Both the source code of Rel-CSKGC and \atomicTT{} are publicly available on \url{https://github.com/NUSTM/Dense-ATOMIC}.
\end{abstract}

\section{Introduction}

\atomic{} is a large-scale human-annotated commonsense knowledge graph focusing on the inferential knowledge in social life~\cite{sap2019atomic}.
It consists of nine \emph{if-then} relation types describing the causes, effects, agent, stative, and theme of an event. 
The research on \atomic{} has drawn more and more attention in recent years. 
An increasing number of downstream tasks, including commonsense reasoning~\cite{DBLP:conf/acl/00020QZ0022}, storytelling~\cite{DBLP:conf/emnlp/BrahmanC20}, question answering~\cite{DBLP:conf/acl/HeoKCZ22}, dialog generation~\cite{DBLP:conf/acl/WuLZW22}, etc., have improved their performances by acquiring and utilizing the commonsense knowledge from \atomic{}.

\begin{figure*}[!htp]
    \centering
    \includegraphics[width=0.99\textwidth]{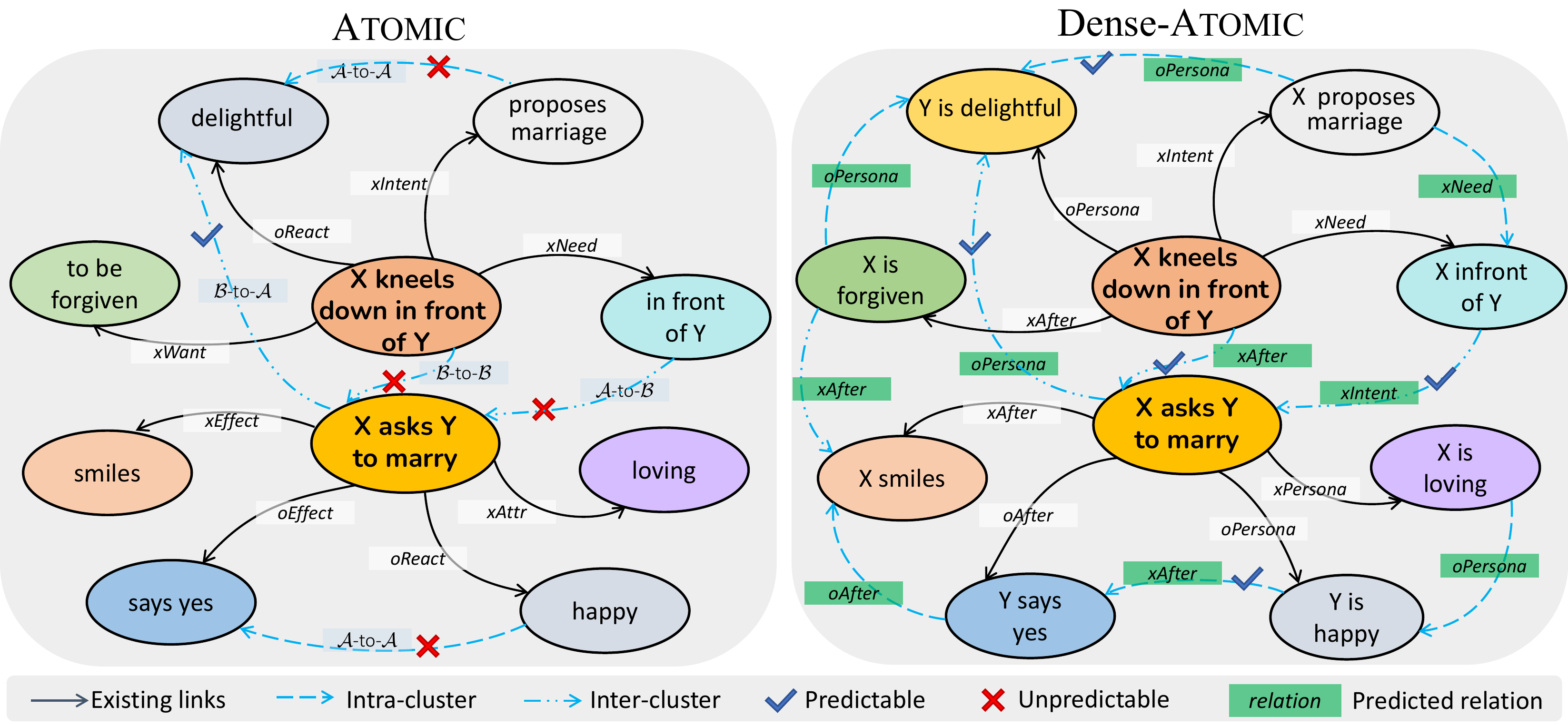}
    \caption{\atomic{} vs. \atomicTT{}. Firstly, \atomicTT{} completes many missing links in \atomic{}, including $\mathcal{B}$-to-$\mathcal{A}$, $\mathcal{B}$-to-$\mathcal{B}$, $\mathcal{A}$-to-$\mathcal{B}$, and $\mathcal{A}$-to-$\mathcal{A}$ links, \emph{e.g.}, missing ``oPersona'' link between \emph{``X proposes marriage''} and \emph{``Y is delightful''} (type: $\mathcal{A}$-to-$\mathcal{A}$); Secondly, \atomicTT{} contains more multi-hop paths, e.g., a two-hop path \emph{``X asks Y to marry''} $\rightarrow$ \emph{``Y says yes''} $\rightarrow$ \emph{``X smiles''} after predicting missing links on normalizd \atomic{}.}
    \label{fig:comparison}
\end{figure*}

Currently, \atomic{} was constructed under one-hop annotations.
It began with 24,000 pre-defined base events and nine relation types.
For each base event and each relation, the annotators were asked to write a possible tail event based on one-hop reasoning.
As shown in Figure~\ref{fig:comparison}, given the base event \emph{``X asks Y to marry''}, the annotated tail events can be \emph{``loving''} under the  relation of \emph{``xAttr''}, \emph{``smiles''}  under the  relation of \emph{``xEffect''}, and \emph{``says yes''} under the relation of \emph{``oEffect''}.

In such a one-hop annotation manner, each base event and its related annotated tail events shape a bipartite graph containing only $\mathcal{B}$-to-$\mathcal{A}$ links, where $\mathcal{B}$ denotes the $\mathbf{B}$ase event and $\mathcal{A}$ denotes the $\mathbf{A}$nnotated tail event.
Thereby, the whole graph of \atomic{} can be viewed as a set of $\mathcal{B}$-to-$\mathcal{A}$ bipartite graphs, while the $\mathcal{B}$-to-$\mathcal{B}$, $\mathcal{A}$-to-$\mathcal{B}$ and $\mathcal{A}$-to-$\mathcal{A}$ links between different bipartite graphs were almost ignored.
In Figure~\ref{fig:comparison}, the dashed lines illustrate such missing links in \atomic{}, e.g., an annotated tail event \emph{``in front of Y''} and a base event \emph{``X asks Y to marry''} in two different bipartite graphs miss a link of the \emph{``xIntent''} relation.

This leads to two shortcomings of \atomic{}.
Firstly, with only $\mathcal{B}$-to-$\mathcal{A}$ links, \atomic{} contains very few multi-hop paths, since an annotated tail event cannot become the \emph{head event} of a triplet.
Secondly, missing $\mathcal{B}$-to-$\mathcal{B}$, $\mathcal{A}$-to-$\mathcal{B}$ and $\mathcal{A}$-to-$\mathcal{A}$ links cause unsatisfactory knowledge coverage, despite its high-quality human-annotated commonsense knowledge.
Both shortcomings limit the potential of \atomic{} in practical applications.
Intuitively, an ideal CSKG requires high knowledge coverage to meet the needs of various tasks, and massive multi-hop paths to understand the evolution between different events.

In this work, we aim to construct a densely-connected \atomic{}.
The key is to complete different types of missing links, leading to denser \atomic{} with high knowledge coverage and massive multi-hop paths.
We achieve this goal through three main steps: Normalizing Tail Events, Training a Relation Prediction Model and Constructing \atomicTT{}.

Firstly, most of the annotated tail events in \atomic{} have different patterns to the base events,
so we normalize annotated tail events in \atomic{} to a consistent pattern (\emph{``Subject + Verb + Object''}), to facilitate subsequent CSKG completion.
Specific relations are also grouped to mitigate ambiguity.

Secondly,
we train a relation prediction model based on a set of existing triplets in \atomic{} to infer the missing links on the whole graph, \emph{i.e.}, CSKG completion upon \atomic{}.
To the best of our knowledge, 
most of the existing studies for CSKG completion 
utilized the translation based methods, which formalized the CSKG completion as a \emph{tail event} ranking task given the \emph{head event} and the relation.
A graph convolutional network (GCN) was mostly employed to encode the graph embeddings of events, but its performance is unsatisfactory since the sparsity of \atomic{} limits the information propagation on the GCN \citep{DBLP:conf/aaai/MalaviyaBBC20}. %\rxia{cite a support paper here}
In contrast, in this work, we propose a method called Rel-CSKGC, which 
regards CSKG completion as a relation prediction problem given the \emph{head event} and the \emph{tail event}, and accordingly train a CSKG completion model based on \atomic{}.

Finally,
based on the CSKG completion model, we construct \atomicTT{} by inferring the missing links on \atomic{}.
Figure~\ref{fig:comparison} illustrates the main differences between \atomic{} and \atomicTT{}.

We conduct extensive evaluations towards the Rel-CSKGC method and the constructed \atomicTT{}, respectively. 

First, we compare Rel-CSKGC with several newly proposed relation prediction methods and translation based methods.
Both automatic evaluation on an annotated subgraph and human evaluation on 500 sampled triplets show the advantage of Rel-CSKGC for completion on \atomic{} .

Next, we evaluate \atomicTT{} from the perspectives of knowledge coverage and multi-hop paths respectively.
Extensive experiments are conducted in terms of statistics, human evaluation, and simple downstream tasks.
The results demonstrate that Dense-ATOMIC surpasses ATOMIC in terms of triplet counts by an order of magnitude, and multi-hop paths by more than two orders of magnitude, respectively, while at the same time maintaining its quality.

\begin{figure*}[!htp]
    \centering
    \includegraphics[width=0.99\textwidth]{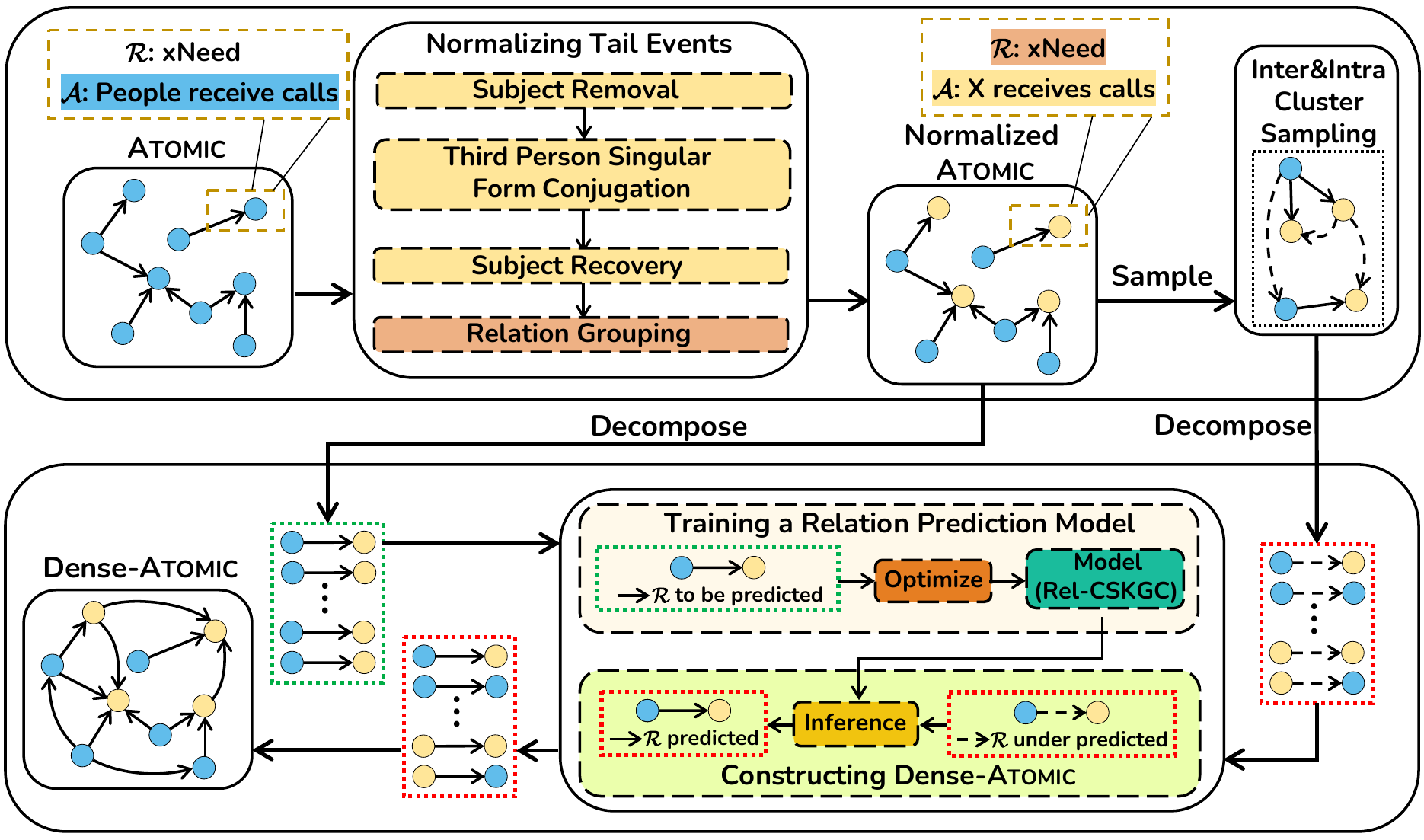}
    \caption{Procedure of constructing \atomicTT{}.}
    \label{fig:procedure}
\end{figure*}

\section{Approach}

Figure~\ref{fig:procedure} illustrates the procedure of constructing \atomicTT{}, consisting of three main steps: Normalizing Tail Events, Training a Relation Prediction Model, and Constructing \atomicTT{}.

\subsection{Normalizing Tail Events}
\label{sec:converting}

\atomic{} contains only $\mathcal{B}$-to-$\mathcal{A}$ triplets.
A CSKG completion model trained with $\mathcal{B}$-to-$\mathcal{A}$ triplets is inapplicable to predict $\mathcal{B}$-to-$\mathcal{B}$, $\mathcal{A}$-to-$\mathcal{A}$, and $\mathcal{A}$-to-$\mathcal{B}$ links, since base events (usually sentences) and annotated tail events (usually phrases or words) have different patterns.
This results in a shortage of knowledge coverage and multi-hop paths during the completion.

To this end, we propose Normalizing Tail Events to convert annotated tail events to the same pattern as the base events, including subject removal, third person singular form conjugation, subject recovery, and relation grouping.

\paragraph{Subject Removal}
For a few annotated tail events being complete sentences, we perform dependency tree parsing and part-of-speech tagging with CoreNLP~\cite{DBLP:conf/acl/ManningSBFBM14} and remove subjects based on the two kinds of structure patterns, which makes the nodes in the graph become a uniform pattern and benefits the subject recovery process.
For example, given a tail event ``He smiles'', we first remove the subject ``He'' and convert it to a universal expression ``Y smiles'' in the subject recovery process.

\paragraph{Third Person Singular Form Conjugation}
In our preliminary experiments, a CSKG completion model tends to correlate phrases starting with \emph{``to''} with relations such as \emph{``xWant''}, \emph{``xIntent''}, so we leverage WordNet~\cite{DBLP:journals/cacm/Miller95} to acquire the verb root and add the suffix (-s, -es, etc.) according to English grammar.

\paragraph{Subject Recovery}
We add subjects to processed annotated tail events based on different relations.

\paragraph{Relation Grouping}
Both \emph{``xWant''} and \emph{``xEffect''} describe the possible subsequent events, distinguished by \emph{``to''} representing subject will.
After third person singular form conjugation, the two relations may lead to ambiguity.
We perform relation grouping for all these relations to mitigate ambiguity.
\emph{``xEffect''} and \emph{``xWant''} form \emph{``xAfter''} describing \emph{what will happen to X}.
\emph{``oEffect''} and \emph{``oWant''} form \emph{``oAfter''} describing \emph{what will happen to Y}.
\emph{``xAttr''} and \emph{``xReact''} form \emph{``xPersona''} describing \emph{how X feels or is described}.
It should be noted that the relation grouping process leads to a non-serious problem, i.e., the grouped relation cannot distinguish between subjective and objective semantics.
However, it mitigates \atomic{}'s sparsity issue and improves the performance of the relation prediction model.

Due to the page limitation, the pseudo-code of normalizing tail events is present in Appendix~\ref{app:node_homogenization}.
 
It is worth noting that our normalization method resembles a prior work~\citep{DBLP:conf/www/FangZWSH21, DBLP:conf/emnlp/FangWCHZSH21}. 
Their purpose is to align \atomic{} with other CSKGs,
while we focus on event alignment in \atomic{} by eliminating differences among different events.

\subsection{Training a Relation Prediction Model}
\label{sec:training_model}

\subsubsection{Limitation of Traditional Methods}

Traditional methods for the completion of \atomic{} proposed to score all candidate \emph{tail events} given the \emph{head event} and the relation.
The GCN for encoding graph embeddings of events induced two shortcomings: 1) it is difficult for a GCN to propagate information due to the sparse graph structure of \atomic{}~\citep{DBLP:conf/aaai/MalaviyaBBC20}; 2) it cannot sufficiently utilize semantic information of events.

\subsubsection{Our Rel-CSKGC Method}

\begin{figure}[!htp]
    \centering
    \includegraphics[width=0.99\columnwidth]{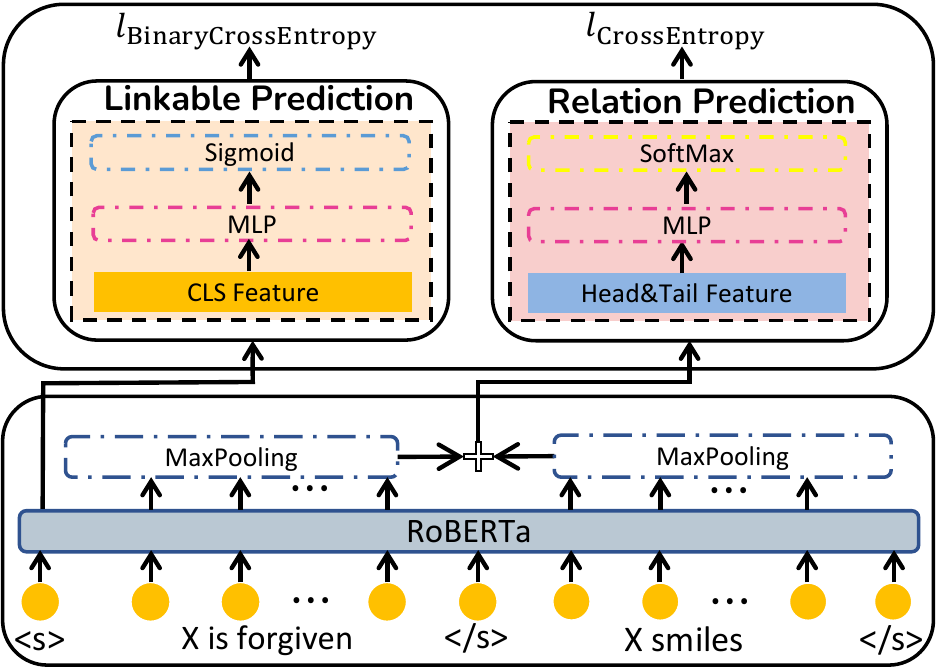}
    \caption{The detailed structure of Rel-CSKGC.}
    \label{fig:model}
\end{figure}

To address these issues, we propose Rel-CSKGC, as illustrated in Figure~\ref{fig:model}.
Specifically, \atomic{} is first decomposed into independent triplets, and then Rel-CSKGC predicts the relation given the \emph{head event} and the \emph{tail event} of a triplet.
Rel-CSKGC utilizes no graph structure information thus avoiding the problem caused by the sparsity.
Additionally, encoding both the \emph{head event} and the \emph{tail event} with the pretrained language model successfully takes advantage of semantic information.

\paragraph{Problem Formulation}
Given a CSKG $G = (N, V)$, where $N$ is the set of nodes and $V$ is the set of edges, we consider a single training instance as a triplet $v_{i} = (h, r, t)$ with the \emph{head event} $h$, \emph{relation type} $r$ and the \emph{tail event} $t$.
Here, $r \in V$ and $h, t \in N$.
The objective of Rel-CSKGC is to predict the most reasonable $r$ given $h$ and $t$.~\footnote{To keep \atomic{} concise, we only predict the most reasonable relation in this work.}

\paragraph{Main Structure}
We utilize RoBERTa~\citep{DBLP:journals/corr/abs-1907-11692} to acquire contextual representations of free-form texts describing events.
The input is the concatenation of $h$ and $t$.
We acquire the embedding matrix of $h$ and $t$ by:
\begin{equation}
    [H;T] = \text{RoBERTa}([h;t])
\end{equation}
where $H \in \mathbb{R}^{|N| \times D}$ and $T \in \mathbb{R}^{|N| \times D}$.
$|N|$ is the number of tokens of the event, and $D$ is the dimensionality of representation.
We apply max pooling on $H$ and $T$ to acquire sentence embeddings $e_h$ and $e_t$. 
The objective function can be defined with trainable weights $W_{t} \in \mathbb{R}^{1 \times D}$ and $W_{c} \in \mathbb{R}^{K \times 2D}$:
\begin{equation}
    o = \text{sigmoid}(W_{t} e_{\texttt{<s>}}) + \text{softmax}(W_{c} (e_h, e_t))
\end{equation}
where $K$ is the number of relations and $e_{\texttt{<s>}}$ the embedding of \texttt{<s>}-token used as a indicator for whether $h$ and $t$ are related.

\paragraph{Negative Sampling}
Rel-CSKGC requires negative samples to predict \emph{unlinkable} links.
We consider the following two strategies to construct negative samples: 1) \textbf{Random} negative sampling.
For a gold triplet, we randomly select an event from normalized \atomic{} as the new \emph{tail event} to replace the original \emph{tail event};
2) \textbf{Persona} negative sampling.
Triplets under relations of \emph{``xPersona''} and \emph{``oPersona''} follow the pattern of \emph{``Subject + is + Adjective''} and account for a large part in \atomic{}.
Models tend to always predict \emph{``xPersona''} or \emph{``oPersona''} when the given tail event follows the pattern of \emph{``Subject + is + Adjective''}.
To alleviate this problem, we specifically construct negative samples by replacing the \emph{tail event} of triplets under relations of \emph{``xPersona''} and \emph{``oPersona''} with a randomly-chosen event containing ``is''.

\subsection{Constructing \atomicTT{}}
\label{sec:inferring}

Based on Rel-CSKGC, we train a relation prediction model with existing triplets in \atomic{} and then use the model to complete missing links in \atomic{}.
We adopt threshold-based link prediction to decide whether two events are related and propose an intra-and-inter cluster completion strategy to reduce the cost of completing entire~\atomic{}.

\paragraph{Threshold-based Link Prediction}
Threshold-based link prediction (TLP) is a heuristic strategy to decide whether a relation is acceptable according to the probability predicted by Rel-CSKGC.
Different thresholds are specifically tuned for different relations.
The model predicts the relation only if the final probability is above the corresponding threshold.
TLP is used in all our models as the last step for the link acceptance decision.

\paragraph{Intra-and-inter Cluster Completion Strategy}
Since it's computationally expensive to iterate over all pairs of \emph{head} and \emph{tail event}s during the inference, we design an intra-and-inter cluster completion strategy to trade off between the completion scale and the time complexity.
In Figure~\ref{fig:comparison}, we consider each base event and its annotated tail events as a \emph{cluster}.
\textbf{Intra-cluster completion} infers missing links inside a cluster.
Intuitively, annotated tail events in one cluster, written based on the same base event, are highly related and may contain more missing links.
\textbf{Inter-cluster completion} infers missing links between different clusters.
Annotated tail events in different clusters are written independently based on different base events, thus links between different clusters are under-explored.

Due to the limited computing resource and time, we temporarily provide the results of 100 sampled clusters in this paper.
Increasing the sampling size can further improve the scale of \atomicTT{}, but that will also linearly increases the computational cost.
We will release versions with larger sampling sizes later.

\section{Evaluation of Our Rel-CSKGC Method}

In this section, we compare Rel-CSKGC with relation prediction and translation based methods by experimenting on a newly annotated subgraph and human evaluation.

\subsection{Training and Test Set Construction}
\label{sec:data}

\paragraph{Training Set with Negative Sampling}
Following~\citet{sap2019atomic}'s split of \atomic{}, we randomly sample negative triplets from the training split with negative sampling strategies introduced in Section~\ref{sec:training_model}.
We combine sampled negative triplets and the training split to construct the training set for Rel-CSKGC.
The statistic of the training set is illustrated in Table~\ref{tab:training_statistics}.~\footnote{The imbalance between random and persona negative sampling methods was established based on a preliminary experiment, which provided insights into optimal sampling sizes.}

\begin{table}[!htp]
    \centering
    % \scriptsize
    % \footnotesize
    \resizebox{1\columnwidth}{!}{
        \begin{tabular}{ccc}
            \hline
            % Base Event & \multicolumn{2}{|c|}{PersonX has a ball} \\ \hline
            \atomic{} & Rand. Neg. Samples & Per. Neg. Samples \\ \hline
            463,264 & 1,890,350 & 756,140 \\ \hline
        \end{tabular}
    }
    \caption{Statistics of the training set for Rel-CSKGC.}
    \label{tab:training_statistics}
\end{table}

\paragraph{Test Set with Annotated Subgraph}
To test the performance of Rel-CSKGC, we construct a ground-truth subgraph by randomly sampling three clusters from the test split and annotating all pairs of \emph{head event}s and \emph{tail event}s with the most reasonable relation.
The statistic of the annotated ground-truth subgraph is shown in Table~\ref{tab:validation_statistics}.

\begin{table}[!htp]
    \centering
    % \small
    % \footnotesize
    % \resizebox{1\columnwidth}{!}{
        \begin{tabular}{cccc}
            \hline
            % Base Event & \multicolumn{2}{|c|}{PersonX has a ball} \\ \hline
            Relation & Total & Intra & Inter \\ \hline
            xAfter & 243 & 186 & 57 \\ 
            xNeed  & 66 & 64 & 2 \\ 
            xIntent & 72 & 51 & 21 \\ 
            xPersona & 291 & 226 & 65 \\ 
            oAfter & 262 & 174 & 88 \\ 
            oPersona & 114 & 70  & 44\\ 
            NoLink & 4234 & 2303 & 1931 \\ \hline
        \end{tabular}
    % }
    \caption{Statistics of the annotated subgraph. Intra and Inter indicate the intra- and inter- cluster, respectively.}
    \label{tab:validation_statistics}
\end{table}

\subsection{Compared Methods}
\label{sec:evaluation}

We select 4 baselines comprising two different types of CSKG completion methods and use the specific evaluation protocol for each of them.

\subsubsection{Relation Prediction Methods}

\paragraph{Baselines}
We adapt \textbf{CE-random}~\citep{li-etal-2016-commonsense}, a method augmenting CSKGs by scoring novel tuples, to predict the missing relation.
We also compare \textbf{KG-BERT}~\citep{DBLP:journals/corr/abs-1909-03193}, which probes the performance of relation prediction methods on knowledge graphs.
Note that we replace BERT~\citep{DBLP:conf/naacl/DevlinCLT19} with RoBERTa~\citep{DBLP:journals/corr/abs-1907-11692} in KG-BERT for fair comparison.

\paragraph{Evaluation Protocal}
Ranking metrics (HITS and Mean Reciprocal Rank) designed for translation based methods are not applicable to relation prediction methods.
By valuing precision more than recall on CSKG completion, we utilize precision for the evaluation of relation prediction methods.

\subsubsection{Translation Based Methods}
\label{sec:trans}

\paragraph{Baselines}
\textbf{SynLink}~\citep{DBLP:conf/aaai/MalaviyaBBC20} proposed to densify the CSKG with synthetic links for better graph representation.
\textbf{InductiveE}~\citep{DBLP:conf/ijcnn/WangWHYLK21} introduced indutive learning on the CSKG by enhancing the unseen event representations with neighboring structure information.

\paragraph{Evaluation Protocal}
To handle the evaluation mismatch between Rel-CSKGC and translation based methods, we designed a transformation strategy.
Specifically, we randomly sample 500 triplets from~\citet{DBLP:conf/aaai/MalaviyaBBC20}'s test split.
For SynLink and InductivE, a threshold is set for hit@1 score, and a \emph{tail event} is accepted only when the score is above the threshold.
We tune the threshold to ensure the number of triplets inferred by Rel-CSKGC, SynLink, and InductivE close on these 500 triplets.
We then calculate the proportion of meaningful triplets for different methods manually.\footnote{In the given context, ``meaningful triplets'' refer to triplets that are considered reasonable, coherent, and non-contradictory by human evaluators.}

\subsection{Main  Results}

\paragraph{Relation Prediction Methods}
In Table~\ref{tab:relkgc}, we compare Rel-CSKGC with different relation prediction methods, and Rel-CSKGC achieves consistent improvement on the test set of the annotated subgraph.
Paired $t$-Test result proves that the improvement of Rel-CSKGC is significant.
From Table~\ref{tab:relkgc}, we can observe that the precision of intra-cluster completion is significantly higher than that of inter-cluster completion for all methods.
This demonstrates that tail events annotated based on the same base event are highly related to each other and easier for models to predict relations, while the prediction for inter-cluster events is more challenging.

\begin{table}[!htp]
    % \resizebox{1\columnwidth}{!}{
    \centering
    % \small
    % \scriptsize
    % \footnotesize
    \begin{tabular}{cccc}
        \hline
        Method & Total & Intra & Inter     \\ 
        \hline
        CE-random & 0.45 & 0.53 & 0.29 \\
        KG-BERT & 0.60  & 0.67 & 0.43 \\
        \hline
        Rel-CSKGC & {\bf 0.68} & {\bf 0.78}  & {\bf 0.51}                 \\
        - w/o random & 0.36 & 0.45 & 0.22        \\
        - w/o persona & 0.58 & 0.66 & 0.44            \\
        \hline
        Rel-CSKGC$_{human}$ & 0.80 & 0.91 & 0.62 \\
        \hline
    \end{tabular}
    % }
    \caption{Rel-CSKGC vs. Relation Prediction methods on Precision. Intra and Inter indicate the result of the intra- and inter- cluster, respectively.}
    \label{tab:relkgc}
\end{table}

\begin{table}[!htp]
    \centering
    \small
    % \resizebox{.99\columnwidth}{!}{
        \begin{tabular}{cccc}
            \hline
            % Base Event & \multicolumn{2}{|c|}{PersonX has a ball} \\ \hline
            Method & \# Predicted & \# Meaningful & Proportion \\ \hline
            SynLink$_{Adapt}$ & 133 & 93 & 0.70 \\ 
            InductivE$_{Adapt}$  & 132 & 106 & 0.80 \\
            Rel-CSKGC & \textbf{174} & \textbf{152} & \textbf{0.87}\\ \hline
        \end{tabular}
    % }
    \caption{Rel-CSKGC vs. Translation Based methods.}
    \label{tab:human_p_evaluation}
\end{table}

\paragraph{Translation Based Methods}
After carefully tuning the threshold based on the strategy in Section~\ref{sec:trans}, Rel-CSKGC, SynLink, and InductivE predict 174, 133, and 132 triplets on 500 randomly sampled triplets.
In Table~\ref{tab:human_p_evaluation}, Rel-CSKGC outperforms SynLink and InductivE by a large margin on proportion and the number of meaningful triplets.

\subsection{Human Evaluation}

\paragraph{Motivation}
Upon observing predictions of Rel-CSKGC, we note that some triplets could be reasonable, while the annotated subgraph doesn't cover them.
For example, given a \emph{head event} ``X accepts Y's apology'' and a \emph{tail event} ``X is generous'', the annotated ground-truth relation is ``xPersona'', while Rel-CSKGC could predict another reasonable relation ``xIntent''.
Consequently, we perform the human evaluation to check whether a predicted triplet is actually meaningful.

\paragraph{Result}
We can find from the last row of Table~\ref{tab:relkgc} that Rel-CSKGC achieves an even higher precision of 0.80, suggesting that Rel-CSKGC can predict reasonable triplets neglected during the subgraph annotation.
The high precision by human evaluation also guarantees the quality of predicted triplets.

\subsection{Ablation Study}

To validate the effectiveness of negative sampling, we report experimental results without negative sampling in Table~\ref{tab:relkgc}.
The performance of Rel-CSKGC drops dramatically without any negative sampling strategies, validating the effectiveness of negative sampling.

\begin{figure}%[!htp]
    \centering
    \resizebox{.99\columnwidth}{!}{
    \includegraphics{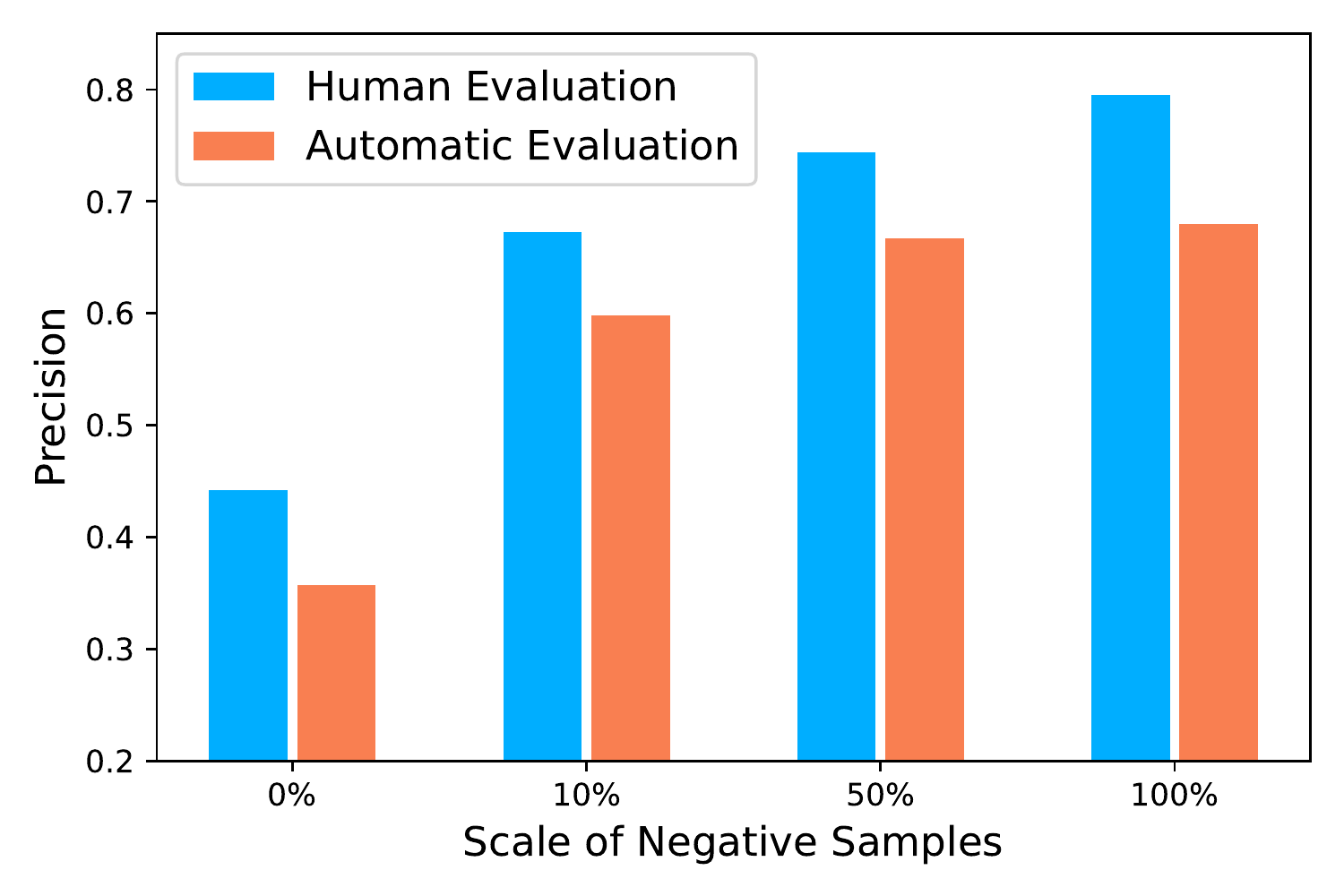}}
    \caption{Precision of Rel-CSKGC with different scales of negative samples on the test set by automatic and human evaluation.}
    \label{fig:ratio}
\end{figure}

By experimenting Rel-CSKGC with different scales of random negative samples in Figure~\ref{fig:ratio}, we find that the precision of Rel-CSKGC increases using both automatic and human evaluation as more negative samples are used for training.

\section{Evaluation of the Constructed \atomicTT{}}

\subsection{Knowledge Coverage and Quality}

In this subsection, we aim to answer the following question: \emph{Does} \atomicTT{} \emph{yield higher knowledge coverage while ensuring the quality?}

To this end, we statistically and manually compare \atomicTT{} with \atomic{} from the following three perspectives.

\begin{table}[!htp]
    \centering
    % \footnotesize
    % \small
    \resizebox{1\columnwidth}{!}{
    \begin{tabular}{crrrr}
        \hline
         & \# Events & \# 1-hop & \# 2-hop & \# 3-hop     \\ 
        \hline
        \atomic{} & 299,068 & 696,321 & 19,231 & 509              \\
        \atomicTT{} & \textbf{283,435}  & \textbf{1,967,373} & \textbf{10,658,242} & \textbf{67,888,373}   \\
        \hline
    \end{tabular}
    }
    \caption{\atomic{} vs. \atomicTT{} on the number of events and multi-hop paths.}
    \label{tab:completion}
\end{table}

\paragraph{\atomicTT{} yields higher knowledge coverage}
In Table~\ref{tab:completion}, we present the comparison between \atomic{} and \atomicTT{}.
\atomicTT{} contains 3x more one-hop paths than \atomic{}, contributing a significantly higher knowledge coverage.
It's worth noting that different tail events in \atomic{} could become the same after normalizing tail events, so \atomicTT{} contains slightly fewer events than \atomic{}.

\paragraph{Triplets in \atomicTT{} have relatively high precision}
In Table~\ref{tab:relkgc}, Rel-CSKGC achieves a precision of 0.80 by human evaluation.
Moreover, from comparison results with translation based methods in Table~\ref{tab:human_p_evaluation}, Rel-CSKGC outperforms two state-of-the-art methods by more than 7 percentage points.
The high performance of Rel-CSKGC ensures the quality of predicted triplets to a certain extent.

\paragraph{\atomicTT{} benefits the performance of \comet{}}
To empirically demonstrate the knowledge coverage and quality of \atomicTT{}, we evaluate \atomicTT{} with \comet{}~\citep{DBLP:conf/acl/BosselutRSMCC19}.
The relation distribution of \atomicTT{} is long-tailed.
We randomly sample 262,678 triplets from predicted triplets and recover the grouped relations to their original relations by following the relation distribution of the \citet{sap2019atomic}'s training split.
Apart from the evaluation of perplexity, we design a strategy to evaluate the diversity score of generated \emph{tail event}s.
For each relation, we randomly sample 10 \emph{head events} from the test set.
For each test sample consisting of a \emph{head event} and a relation, 10 candidates are generated using beam search.
For each candidate, we manually give a score of 0, 1, or 2, representing ``unreasonable'', ``plausible'', and ``reasonable'', respectively.
We then merge candidates of similar semantics into a group and calculate the group average score. 
The diversity score of 10 candidates is the sum of the group scores.
Intuitively, the lower perplexity and the higher diversity score indicate the higher knowledge quality and the higher knowledge coverage of \atomicTT{}, and \cometTT{} outperforms \comet{} on both metrics in Table~\ref{tab:relkgc_comet}.
In Table~\ref{tab:atomic_example_cp}, we can find that tail events generated by \cometTT{} are more semantically different.

\begin{table}%[!htp]
            \centering
            % \small
            % \resizebox{.99\columnwidth}{!}{
            \begin{tabular}{crr}
                \hline
                 & PPL~$\downarrow$ & DS~$\uparrow$     \\ 
                \hline
                \comet{} & 11.14 & 9.16                  \\
                \cometTT{} &{\bf 11.11} & {\bf 10.77}                  \\
                \hline
            \end{tabular}
            % }
            \caption{\comet{} vs. \cometTT{}. PPL and DS indicate perplexity and diversity score, respectively.}
            \label{tab:relkgc_comet}
\end{table}

\subsection{Multi-hop Paths in \atomicTT{}}

\begin{table}%[!htp]
            \centering
            % \small
            % \scriptsize
            % \resizebox{.99\columnwidth}{!}{
                \begin{tabular}{cc}
                    \hline
                    % Base Event & \multicolumn{2}{|c|}{PersonX has a ball} \\ \hline
                    {\bf \comet{}} & {\bf \cometTT{}} \\ \hline
                    \textcolor[RGB]{255, 0, 191}{\textbf{to study hard}} & \textcolor[RGB]{255, 0, 191}{\textbf{to study harder}} \\
                    \textcolor[RGB]{255, 0, 191}{\textbf{study hard}} & \textcolor[RGB]{255, 0, 191}{\textbf{to study more}} \\ 
                    \textcolor[RGB]{255, 0, 191}{\textbf{to study more}} & \textcolor[RGB]{19, 208, 171}{\textbf{to get a good grade}} \\ 
                    \textcolor[RGB]{255, 0, 191}{\textbf{to study}} & \textbf{to take a test}  \\ 
                    \textcolor[RGB]{19, 208, 171}{\textbf{to get a good grade}} &  \textcolor{orange}{\textbf{to do well in school}} \\ 
                    \textbf{to take a test} &  \textcolor{orange}{\textbf{to do well in class}} \\ 
                    \textcolor{orange}{\textbf{to do well in school}} & \textcolor[RGB]{0, 119, 200}{\textbf{to apply for a job}} \\ 
                    \textcolor[RGB]{0, 119, 200}{\textbf{to get a good job}} & \textcolor[RGB]{120, 163, 0}{\textbf{to pass the class}} \\ 
                    \textcolor[RGB]{0, 119, 200}{\textbf{to apply for a job}} & \textcolor[RGB]{41, 196, 208}{\textbf{to get a prize}} \\ 
                    \textcolor[RGB]{0, 119, 200}{\textbf{to apply for a good job}} & \textcolor{purple}{\textbf{to go to school}} \\ \hline
                \end{tabular}
            % }
            \caption{Events generated by \comet{} and \cometTT{} given ``\textbf{\emph{X needs a good grade}}'' and ``\textbf{\emph{xWant}}''. Semantically similar events are in the same color.}
            % The same color means the similar semantic and the same group.
            \label{tab:atomic_example_cp}
\end{table}

\begin{table}%[!htp]
    % \resizebox{1\columnwidth}{!}{
    \centering
    % \small
    % \scriptsize
    % \footnotesize
    \begin{tabular}{cccc}
        \hline
        Sampling Method & 2-hop & 3-hop & 4-hop     \\ \hline
        Random & 0.69 & 0.62  & 0.50 \\
        Heuristic Rule & {\bf 0.84} & {\bf 0.77} & {\bf 0.74}        \\
        \hline
    \end{tabular}
    % }
    \caption{Random vs. Heuristic Rule on human evaluation of sampled multi-hop paths.}
    \label{tab:comparison_reasoning}
\end{table}

\begin{table*}%[!htp]
        \centering
        \small
        % \resizebox{2\columnwidth}{!}{
            \begin{tabular}{c}
            \hline
            \textbf{\textit{2-hop paths}}\\
            \hline

            X misses Y's opportunity $\xrightarrow{xAfter}$ X goes home sadly $\xrightarrow{xPersona}$ X is melancholy \\[0.3ex]

            X takes advantage of the opportunities $\xrightarrow{xAfter}$ X contines to succeed $\xrightarrow{oPersona}$ Y is satisfied \\[0.3ex]
            
            X goes back home $\xrightarrow{xAfter}$ X becomes sleepy $\xrightarrow{xAfter}$ X goes back to his own bed \\[0.3ex]

            X reaches X's goal $\xrightarrow{xAfter}$ X gets an award $\xrightarrow{oAfter}$ Y celebrates their win \\[0.3ex]
 
            \hline
            \textbf{\textit{3-hop paths}}\\
            \hline

            X returns to X's work $\xrightarrow{xAfter}$ X goes home for the day $\xrightarrow{xAfter}$ X sleeps at night $\xrightarrow{oAfter}$ Y is glad to see X slept normally \\[0.3ex]
            
            X plays a role in the development $\xrightarrow{xAfter}$ X receives an award $\xrightarrow{xAfter}$ X gets compliments $\xrightarrow{xAfter}$ X smiles \\[0.3ex]
            
            X talkes about X's feeling $\xrightarrow{xAfter}$ X starts crying $\xrightarrow{xAfter}$ X wipes the tears $\xrightarrow{xPersona}$ X is thankful \\[0.3ex]

            X improves X's chances $\xrightarrow{xAfter}$ X wins the game $\xrightarrow{xAfter}$ X jumps up and down with joy $\xrightarrow{oPersona} $ Y is pleased \\[0.3ex]
            
            \hline
            \end{tabular}
        % }
        \caption{Examples of multi-hop paths randomly sampled from \atomicTT{}.}
        \label{tab:main_case_study}
\end{table*}

The aim of this subsection is to answer the question: \emph{Can multi-hop paths in \atomicTT{} better present the commonsense knowledge?}

Accordingly, we evaluate multi-hop paths based on the human evaluation and performing a newly designed Commonsense Reasoning experiment, respectively: 

\paragraph{Human evaluation confirms the correctness of multi-hop paths in \atomicTT{}} % 逻辑正确性
In Table~\ref{tab:completion}, we have already shown that \atomicTT{} contains orders of magnitude more two-hop and three-hop paths than \atomic{}.
Now, to further validate the correctness of multi-hop paths, we perform the human evaluation on sampled paths to calculate the proportion of reasonable paths.
Note that it's a common phenomenon (both KGs and CSKGs) that \emph{A $\rightarrow$ B} and \emph{B $\rightarrow$ C} are reasonable, while \emph{A $\rightarrow$ B $\rightarrow$ C} is irrational.
For example, \{\emph{Beethoven}, \emph{owner}, \emph{piano}\} and \{\emph{piano}, \emph{color}, \emph{black}\} are two reasonable triplets, but ``\emph{Beethoven}'' and ``\emph{black}'' are not related.
Consequently, we additionally design a simple heuristic sampling rule: a multi-hop path \emph{A $\rightarrow$ \ldots $\rightarrow$ C} is chosen only when A and C are also linked in \atomicTT{}.
By comparing with random sampling in Table~\ref{tab:comparison_reasoning}, we can find that heuristic rule sampling consistently outperforms random sampling: the longer the multi-hop paths, the more significant the improvement.
Multi-hop paths randomly sampled from \atomicTT{} with two different methods are illustrated in Table~\ref{tab:main_case_study}.

\paragraph{\atomicTT{} has the potential of providing contextual information for Commonsense Reasoning}
In order to further validate the effectiveness of multi-hop paths in \atomicTT{}, we utilize BART~\citep{DBLP:conf/acl/LewisLGGMLSZ20} to perform generative Commonsense Reasoning with or without multi-hop paths.
Specifically, with the heuristic rule above, we randomly sample 5000 four-hop paths from \atomicTT{} as the training samples.
For test samples, we manually select 500 reasonable paths from \atomicTT{}. 
% The same sampling method is adopted for test samples except that unreasonable paths are filtered manually for high quality, and we stop sampling when the number of test samples reaches 500.
BART is trained to generate the subsequent event in two different settings: 1) given only the first node of the path; 2) given the first four nodes of the path.
From Table~\ref{tab:comparison_multihop}, we can find that BART trained with multi-hop paths achieves better performance in that multi-hop paths could provide more contextual information useful for Commonsense Reasoning.

\begin{table}[!htp]
    % \resizebox{1\columnwidth}{!}{
    \centering
    % \small
    % \scriptsize
    % \footnotesize
    \begin{tabular}{cccc}
        \hline
         & Bleu-1 & Bleu-2 & ROUGE-L  \\ \hline
        One-hop & 48.57 & 14.24  & 35.58 \\
        Multi-hop & \textbf{48.63} & \textbf{14.93} & \textbf{36.90} \\
        \hline
    \end{tabular}
    % }
    \caption{Scores of tail events generated with one-hop and multi-hop paths.}
    % \caption{The quality of BART-generated subsequent with One-hop and Multi-hop paths.}
    % \caption{Comparison of the quality of subsequent event generated by BART. Each approach uses greedy decoding.}
    \label{tab:comparison_multihop}
\end{table}

\section{Related Work}

ConceptNet~\cite{DBLP:conf/aaai/SpeerCH17} is a large-scale CSKG merging various knowledge bases.
ASER~\cite{DBLP:conf/www/ZhangLPSL20} contains the selectional preference knowledge extracted from more than 11 billion-token unstructured textual data.
TransOMCS~\cite{DBLP:conf/ijcai/ZhangKSR20} utilizes linguistic graphs to convert ASER into the same representation as ConceptNet.
DISCOS~\cite{DBLP:conf/www/FangZWSH21} aggregates the neighboring information to distill the commonsense knowledge in ASER.

Recent years have seen crowdsourced CSKGs aiming to provide high-quality commonsense knowledge triplets.
\citet{sap2019atomic} released \atomic{} consisting of if-then knowledge triplets mainly about daily events.
\citet{DBLP:conf/aaai/HwangBBDSBC21} augmented \atomic{} with event-centered and physical-entity triplets.
GLUCOSE~\cite{DBLP:conf/emnlp/MostafazadehKMB20} grounds the implicit commonsense knowledge about everyday situations in a narrative context for richer inferential content.

\atomicTT{} unleashes the power of \atomic{} for high knowledge coverage and multi-hop paths.

% Similar to conventional KGs, CSKGs also have missing links.
Prior CSKG completion methods performed binary classification by scoring BiLSTM-encoded tuples~\cite{li-etal-2016-commonsense, saito-etal-2018-commonsense, jastrzebski-etal-2018-commonsense}.
Following translation based methods for the knowledge graph completion~\citep{DBLP:conf/aaai/DettmersMS018, DBLP:conf/aaai/ShangTHBHZ19, DBLP:conf/ijcai/MeilickeCRS19, DBLP:conf/iclr/QuCXBT21, DBLP:conf/emnlp/ZhangLJLWJY21, DBLP:conf/acl/LovelaceNVLR20},~\citet{DBLP:conf/aaai/MalaviyaBBC20} additionally densified the CSKG based on BERT similarity and achieve promising results.
\citet{DBLP:conf/ijcnn/WangWHYLK21} and \citet{DBLP:conf/cikm/JuYL22} designed heuristic rules to add more edges for nodes with fewer neighbors.
\citet{DBLP:conf/acl/MoghimifarQZHB20} presented a neural-symbolic reasoner to learn logic rules during the training, making the CSKG completion process interpretable.

Rel-CSKGC differs from them in that we utilize pretrained language models to predict the relation given the \emph{head event} and the \emph{tail event}.
Similar relation prediction methods targeting at the knowledge graph completion have been proposed~\cite{DBLP:conf/nips/SocherCMN13, DBLP:journals/corr/abs-1909-03193, 10.1145/3366423.3380123}.
To our best knowledge, we are the first to explore the relation prediction method on CSKG completion.

\section{Conclusion}

In this paper, we construct \atomicTT{} for high knowledge coverage and massive multi-hop paths and accordingly propose a CSKG completion method called Rel-CSKGC to train a relation prediction model and infer the missing links in \atomic{}.
Both automatic and human evaluation show the advantage of Rel-CSKGC over strong baselines.
The statistics prove that \atomicTT{} has significantly more triplets and multi-hop paths, providing potential for high-quality downstream applications and multi-hop reasoning based on commonsense knowledge.

\section*{Limitations}

Our approach for constructing \atomicTT{} still has two limitations:
1) to keep \atomicTT{} simple, we only consider the most reasonable relation in this paper, while the relation between two events can be complex and diversified.
We will release versions of \atomicTT{} with diversified relations later;
2) due to page limitation, we only evaluate \atomicTT{} on simple commonsense reasoning tasks, and we will further validate the multi-hop reasoning capacity of \atomicTT{} on more complex downstream tasks in the future.

\section*{Ethics Statement}

We would like to thank the Allen Institute for AI for their valuable work on \atomic{}.
The \atomic{} is licensed under a license of CC BY, which allows remixing, transforming, and building upon the material for any purpose.
We will also make our \atomicTT{} publicly available later.
\citet{DBLP:conf/emnlp/MehrabiZMPRG21} have found representational harms in common sense resources.
We acknowledge that the generated commonsense from our models might contain biases.
All of the datasets and models are in English, which benefits English speakers more.
We have employed 3 postgraduates experienced in natural language processing for annotation and human evaluation.
We pay postgraduates around \$8 per hour, well above the local average wage, and engage in constructive discussions if they have concerns about the process.

\section*{Acknowledgments}

This work was supported by the Natural Science Foundation of China (No. 62076133), and the Natural Science Foundation of Jiangsu Province for Distinguished Young Scholars (No. BK20200018).

\bibliography{my_bib}

\begin{thebibliography}{35}
\expandafter\ifx\csname natexlab\endcsname\relax\def\natexlab#1{#1}\fi

\bibitem[{Bosselut et~al.(2019)Bosselut, Rashkin, Sap, Malaviya, Celikyilmaz,
  and Choi}]{DBLP:conf/acl/BosselutRSMCC19}
Antoine Bosselut, Hannah Rashkin, Maarten Sap, Chaitanya Malaviya, Asli
  Celikyilmaz, and Yejin Choi. 2019.
\newblock \href {https://doi.org/10.18653/v1/p19-1470} {{COMET:} commonsense
  transformers for automatic knowledge graph construction}.
\newblock In \emph{Proceedings of the 57th Conference of the Association for
  Computational Linguistics, {ACL} 2019, Florence, Italy, July 28- August 2,
  2019, Volume 1: Long Papers}, pages 4762--4779. Association for Computational
  Linguistics.

\bibitem[{Brahman and Chaturvedi(2020)}]{DBLP:conf/emnlp/BrahmanC20}
Faeze Brahman and Snigdha Chaturvedi. 2020.
\newblock \href {https://doi.org/10.18653/v1/2020.emnlp-main.426} {Modeling
  protagonist emotions for emotion-aware storytelling}.
\newblock In \emph{Proceedings of the 2020 Conference on Empirical Methods in
  Natural Language Processing, {EMNLP} 2020, Online, November 16-20, 2020},
  pages 5277--5294. Association for Computational Linguistics.

\bibitem[{Cao et~al.(2020)Cao, Wang, Huang, and Hu}]{10.1145/3366423.3380123}
Ermei Cao, Difeng Wang, Jiacheng Huang, and Wei Hu. 2020.
\newblock \href {https://doi.org/10.1145/3366423.3380123} {\emph{Open Knowledge
  Enrichment for Long-Tail Entities}}, page 384–394. Association for
  Computing Machinery, New York, NY, USA.

\bibitem[{Dettmers et~al.(2018)Dettmers, Minervini, Stenetorp, and
  Riedel}]{DBLP:conf/aaai/DettmersMS018}
Tim Dettmers, Pasquale Minervini, Pontus Stenetorp, and Sebastian Riedel. 2018.
\newblock \href
  {https://www.aaai.org/ocs/index.php/AAAI/AAAI18/paper/view/17366}
  {Convolutional 2d knowledge graph embeddings}.
\newblock In \emph{Proceedings of the Thirty-Second {AAAI} Conference on
  Artificial Intelligence, (AAAI-18), the 30th innovative Applications of
  Artificial Intelligence (IAAI-18), and the 8th {AAAI} Symposium on
  Educational Advances in Artificial Intelligence (EAAI-18), New Orleans,
  Louisiana, USA, February 2-7, 2018}, pages 1811--1818. {AAAI} Press.

\bibitem[{Devlin et~al.(2019)Devlin, Chang, Lee, and
  Toutanova}]{DBLP:conf/naacl/DevlinCLT19}
Jacob Devlin, Ming{-}Wei Chang, Kenton Lee, and Kristina Toutanova. 2019.
\newblock \href {https://doi.org/10.18653/v1/n19-1423} {{BERT:} pre-training of
  deep bidirectional transformers for language understanding}.
\newblock In \emph{Proceedings of the 2019 Conference of the North American
  Chapter of the Association for Computational Linguistics: Human Language
  Technologies, {NAACL-HLT} 2019, Minneapolis, MN, USA, June 2-7, 2019, Volume
  1 (Long and Short Papers)}, pages 4171--4186. Association for Computational
  Linguistics.

\bibitem[{Fang et~al.(2021{\natexlab{a}})Fang, Wang, Choi, Hao, Zhang, Song,
  and He}]{DBLP:conf/emnlp/FangWCHZSH21}
Tianqing Fang, Weiqi Wang, Sehyun Choi, Shibo Hao, Hongming Zhang, Yangqiu
  Song, and Bin He. 2021{\natexlab{a}}.
\newblock \href {https://doi.org/10.18653/v1/2021.emnlp-main.705} {Benchmarking
  commonsense knowledge base population with an effective evaluation dataset}.
\newblock In \emph{Proceedings of the 2021 Conference on Empirical Methods in
  Natural Language Processing, {EMNLP} 2021, Virtual Event / Punta Cana,
  Dominican Republic, 7-11 November, 2021}, pages 8949--8964. Association for
  Computational Linguistics.

\bibitem[{Fang et~al.(2021{\natexlab{b}})Fang, Zhang, Wang, Song, and
  He}]{DBLP:conf/www/FangZWSH21}
Tianqing Fang, Hongming Zhang, Weiqi Wang, Yangqiu Song, and Bin He.
  2021{\natexlab{b}}.
\newblock \href {https://doi.org/10.1145/3442381.3450117} {{DISCOS:} bridging
  the gap between discourse knowledge and commonsense knowledge}.
\newblock In \emph{{WWW} '21: The Web Conference 2021, Virtual Event /
  Ljubljana, Slovenia, April 19-23, 2021}, pages 2648--2659. {ACM} / {IW3C2}.

\bibitem[{Heo et~al.(2022)Heo, Kim, Choi, and Zhang}]{DBLP:conf/acl/HeoKCZ22}
Yu{-}Jung Heo, Eun{-}Sol Kim, Woo~Suk Choi, and Byoung{-}Tak Zhang. 2022.
\newblock \href {https://doi.org/10.18653/v1/2022.acl-long.29} {Hypergraph
  transformer: Weakly-supervised multi-hop reasoning for knowledge-based visual
  question answering}.
\newblock In \emph{Proceedings of the 60th Annual Meeting of the Association
  for Computational Linguistics (Volume 1: Long Papers), {ACL} 2022, Dublin,
  Ireland, May 22-27, 2022}, pages 373--390. Association for Computational
  Linguistics.

\bibitem[{Hwang et~al.(2021)Hwang, Bhagavatula, Bras, Da, Sakaguchi, Bosselut,
  and Choi}]{DBLP:conf/aaai/HwangBBDSBC21}
Jena~D. Hwang, Chandra Bhagavatula, Ronan~Le Bras, Jeff Da, Keisuke Sakaguchi,
  Antoine Bosselut, and Yejin Choi. 2021.
\newblock \href {https://ojs.aaai.org/index.php/AAAI/article/view/16792}
  {(comet-) atomic 2020: On symbolic and neural commonsense knowledge graphs}.
\newblock In \emph{Thirty-Fifth {AAAI} Conference on Artificial Intelligence,
  {AAAI} 2021, Thirty-Third Conference on Innovative Applications of Artificial
  Intelligence, {IAAI} 2021, The Eleventh Symposium on Educational Advances in
  Artificial Intelligence, {EAAI} 2021, Virtual Event, February 2-9, 2021},
  pages 6384--6392. {AAAI} Press.

\bibitem[{Jastrz{\k{e}}bski et~al.(2018)Jastrz{\k{e}}bski, Bahdanau, Hosseini,
  Noukhovitch, Bengio, and Cheung}]{jastrzebski-etal-2018-commonsense}
Stanislaw Jastrz{\k{e}}bski, Dzmitry Bahdanau, Seyedarian Hosseini, Michael
  Noukhovitch, Yoshua Bengio, and Jackie Cheung. 2018.
\newblock \href {https://doi.org/10.18653/v1/W18-1002} {Commonsense mining as
  knowledge base completion? a study on the impact of novelty}.
\newblock In \emph{Proceedings of the Workshop on Generalization in the Age of
  Deep Learning}, pages 8--16, New Orleans, Louisiana. Association for
  Computational Linguistics.

\bibitem[{Ju et~al.(2022)Ju, Yang, and Liu}]{DBLP:conf/cikm/JuYL22}
Jinhao Ju, Deqing Yang, and Jingping Liu. 2022.
\newblock \href {https://doi.org/10.1145/3511808.3557564} {Commonsense
  knowledge base completion with relational graph attention network and
  pre-trained language model}.
\newblock In \emph{Proceedings of the 31st {ACM} International Conference on
  Information {\&} Knowledge Management, Atlanta, GA, USA, October 17-21,
  2022}, pages 4104--4108. {ACM}.

\bibitem[{Lewis et~al.(2020)Lewis, Liu, Goyal, Ghazvininejad, Mohamed, Levy,
  Stoyanov, and Zettlemoyer}]{DBLP:conf/acl/LewisLGGMLSZ20}
Mike Lewis, Yinhan Liu, Naman Goyal, Marjan Ghazvininejad, Abdelrahman Mohamed,
  Omer Levy, Veselin Stoyanov, and Luke Zettlemoyer. 2020.
\newblock \href {https://doi.org/10.18653/v1/2020.acl-main.703} {{BART:}
  denoising sequence-to-sequence pre-training for natural language generation,
  translation, and comprehension}.
\newblock In \emph{Proceedings of the 58th Annual Meeting of the Association
  for Computational Linguistics, {ACL} 2020, Online, July 5-10, 2020}, pages
  7871--7880. Association for Computational Linguistics.

\bibitem[{Li et~al.(2016)Li, Taheri, Tu, and Gimpel}]{li-etal-2016-commonsense}
Xiang Li, Aynaz Taheri, Lifu Tu, and Kevin Gimpel. 2016.
\newblock \href {https://doi.org/10.18653/v1/P16-1137} {Commonsense knowledge
  base completion}.
\newblock In \emph{Proceedings of the 54th Annual Meeting of the Association
  for Computational Linguistics (Volume 1: Long Papers)}, pages 1445--1455,
  Berlin, Germany. Association for Computational Linguistics.

\bibitem[{Liu et~al.(2019)Liu, Ott, Goyal, Du, Joshi, Chen, Levy, Lewis,
  Zettlemoyer, and Stoyanov}]{DBLP:journals/corr/abs-1907-11692}
Yinhan Liu, Myle Ott, Naman Goyal, Jingfei Du, Mandar Joshi, Danqi Chen, Omer
  Levy, Mike Lewis, Luke Zettlemoyer, and Veselin Stoyanov. 2019.
\newblock \href {http://arxiv.org/abs/1907.11692} {Roberta: {A} robustly
  optimized {BERT} pretraining approach}.
\newblock \emph{CoRR}, abs/1907.11692.

\bibitem[{Lovelace et~al.(2021)Lovelace, Newman{-}Griffis, Vashishth, Lehman,
  and Ros{\'{e}}}]{DBLP:conf/acl/LovelaceNVLR20}
Justin Lovelace, Denis Newman{-}Griffis, Shikhar Vashishth, Jill~Fain Lehman,
  and Carolyn~P. Ros{\'{e}}. 2021.
\newblock \href {https://doi.org/10.18653/v1/2021.acl-long.82} {Robust
  knowledge graph completion with stacked convolutions and a student re-ranking
  network}.
\newblock In \emph{Proceedings of the 59th Annual Meeting of the Association
  for Computational Linguistics and the 11th International Joint Conference on
  Natural Language Processing, {ACL/IJCNLP} 2021, (Volume 1: Long Papers),
  Virtual Event, August 1-6, 2021}, pages 1016--1029. Association for
  Computational Linguistics.

\bibitem[{Malaviya et~al.(2020)Malaviya, Bhagavatula, Bosselut, and
  Choi}]{DBLP:conf/aaai/MalaviyaBBC20}
Chaitanya Malaviya, Chandra Bhagavatula, Antoine Bosselut, and Yejin Choi.
  2020.
\newblock \href {https://ojs.aaai.org/index.php/AAAI/article/view/5684}
  {Commonsense knowledge base completion with structural and semantic context}.
\newblock In \emph{The Thirty-Fourth {AAAI} Conference on Artificial
  Intelligence, {AAAI} 2020, The Thirty-Second Innovative Applications of
  Artificial Intelligence Conference, {IAAI} 2020, The Tenth {AAAI} Symposium
  on Educational Advances in Artificial Intelligence, {EAAI} 2020, New York,
  NY, USA, February 7-12, 2020}, pages 2925--2933. {AAAI} Press.

\bibitem[{Manning et~al.(2014)Manning, Surdeanu, Bauer, Finkel, Bethard, and
  McClosky}]{DBLP:conf/acl/ManningSBFBM14}
Christopher~D. Manning, Mihai Surdeanu, John Bauer, Jenny~Rose Finkel, Steven
  Bethard, and David McClosky. 2014.
\newblock \href {https://doi.org/10.3115/v1/p14-5010} {The stanford corenlp
  natural language processing toolkit}.
\newblock In \emph{Proceedings of the 52nd Annual Meeting of the Association
  for Computational Linguistics, {ACL} 2014, June 22-27, 2014, Baltimore, MD,
  USA, System Demonstrations}, pages 55--60. The Association for Computer
  Linguistics.

\bibitem[{Mehrabi et~al.(2021)Mehrabi, Zhou, Morstatter, Pujara, Ren, and
  Galstyan}]{DBLP:conf/emnlp/MehrabiZMPRG21}
Ninareh Mehrabi, Pei Zhou, Fred Morstatter, Jay Pujara, Xiang Ren, and Aram
  Galstyan. 2021.
\newblock \href {https://doi.org/10.18653/v1/2021.emnlp-main.410} {Lawyers are
  dishonest? quantifying representational harms in commonsense knowledge
  resources}.
\newblock In \emph{Proceedings of the 2021 Conference on Empirical Methods in
  Natural Language Processing, {EMNLP} 2021, Virtual Event / Punta Cana,
  Dominican Republic, 7-11 November, 2021}, pages 5016--5033. Association for
  Computational Linguistics.

\bibitem[{Meilicke et~al.(2019)Meilicke, Chekol, Ruffinelli, and
  Stuckenschmidt}]{DBLP:conf/ijcai/MeilickeCRS19}
Christian Meilicke, Melisachew~Wudage Chekol, Daniel Ruffinelli, and Heiner
  Stuckenschmidt. 2019.
\newblock \href {https://doi.org/10.24963/ijcai.2019/435} {Anytime bottom-up
  rule learning for knowledge graph completion}.
\newblock In \emph{Proceedings of the Twenty-Eighth International Joint
  Conference on Artificial Intelligence, {IJCAI} 2019, Macao, China, August
  10-16, 2019}, pages 3137--3143. ijcai.org.

\bibitem[{Miller(1995)}]{DBLP:journals/cacm/Miller95}
George~A. Miller. 1995.
\newblock \href {https://doi.org/10.1145/219717.219748} {Wordnet: {A} lexical
  database for english}.
\newblock \emph{Commun. {ACM}}, 38(11):39--41.

\bibitem[{Moghimifar et~al.(2021)Moghimifar, Qu, Zhuo, Haffari, and
  Baktashmotlagh}]{DBLP:conf/acl/MoghimifarQZHB20}
Farhad Moghimifar, Lizhen Qu, Terry~Yue Zhuo, Gholamreza Haffari, and Mahsa
  Baktashmotlagh. 2021.
\newblock \href {https://doi.org/10.18653/v1/2021.acl-short.100}
  {Neural-symbolic commonsense reasoner with relation predictors}.
\newblock In \emph{Proceedings of the 59th Annual Meeting of the Association
  for Computational Linguistics and the 11th International Joint Conference on
  Natural Language Processing, {ACL/IJCNLP} 2021, (Volume 2: Short Papers),
  Virtual Event, August 1-6, 2021}, pages 797--802. Association for
  Computational Linguistics.

\bibitem[{Mostafazadeh et~al.(2020)Mostafazadeh, Kalyanpur, Moon, Buchanan,
  Berkowitz, Biran, and Chu{-}Carroll}]{DBLP:conf/emnlp/MostafazadehKMB20}
Nasrin Mostafazadeh, Aditya Kalyanpur, Lori Moon, David~W. Buchanan, Lauren
  Berkowitz, Or~Biran, and Jennifer Chu{-}Carroll. 2020.
\newblock \href {https://doi.org/10.18653/v1/2020.emnlp-main.370} {{GLUCOSE:}
  generalized and contextualized story explanations}.
\newblock In \emph{Proceedings of the 2020 Conference on Empirical Methods in
  Natural Language Processing, {EMNLP} 2020, Online, November 16-20, 2020},
  pages 4569--4586. Association for Computational Linguistics.

\bibitem[{Qu et~al.(2021)Qu, Chen, Xhonneux, Bengio, and
  Tang}]{DBLP:conf/iclr/QuCXBT21}
Meng Qu, Junkun Chen, Louis{-}Pascal A.~C. Xhonneux, Yoshua Bengio, and Jian
  Tang. 2021.
\newblock \href {https://openreview.net/forum?id=tGZu6DlbreV} {Rnnlogic:
  Learning logic rules for reasoning on knowledge graphs}.
\newblock In \emph{9th International Conference on Learning Representations,
  {ICLR} 2021, Virtual Event, Austria, May 3-7, 2021}. OpenReview.net.

\bibitem[{Saito et~al.(2018)Saito, Nishida, Asano, and
  Tomita}]{saito-etal-2018-commonsense}
Itsumi Saito, Kyosuke Nishida, Hisako Asano, and Junji Tomita. 2018.
\newblock \href {https://doi.org/10.18653/v1/K18-1014} {Commonsense knowledge
  base completion and generation}.
\newblock In \emph{Proceedings of the 22nd Conference on Computational Natural
  Language Learning}, pages 141--150, Brussels, Belgium. Association for
  Computational Linguistics.

\bibitem[{Sap et~al.(2019)Sap, Le~Bras, Allaway, Bhagavatula, Lourie, Rashkin,
  Roof, Smith, and Choi}]{sap2019atomic}
Maarten Sap, Ronan Le~Bras, Emily Allaway, Chandra Bhagavatula, Nicholas
  Lourie, Hannah Rashkin, Brendan Roof, Noah~A Smith, and Yejin Choi. 2019.
\newblock Atomic: An atlas of machine commonsense for if-then reasoning.
\newblock In \emph{Proceedings of the AAAI Conference on Artificial
  Intelligence}, 01, pages 3027--3035.

\bibitem[{Shang et~al.(2019)Shang, Tang, Huang, Bi, He, and
  Zhou}]{DBLP:conf/aaai/ShangTHBHZ19}
Chao Shang, Yun Tang, Jing Huang, Jinbo Bi, Xiaodong He, and Bowen Zhou. 2019.
\newblock \href {https://doi.org/10.1609/aaai.v33i01.33013060} {End-to-end
  structure-aware convolutional networks for knowledge base completion}.
\newblock In \emph{The Thirty-Third {AAAI} Conference on Artificial
  Intelligence, {AAAI} 2019, The Thirty-First Innovative Applications of
  Artificial Intelligence Conference, {IAAI} 2019, The Ninth {AAAI} Symposium
  on Educational Advances in Artificial Intelligence, {EAAI} 2019, Honolulu,
  Hawaii, USA, January 27 - February 1, 2019}, pages 3060--3067. {AAAI} Press.

\bibitem[{Socher et~al.(2013)Socher, Chen, Manning, and
  Ng}]{DBLP:conf/nips/SocherCMN13}
Richard Socher, Danqi Chen, Christopher~D. Manning, and Andrew~Y. Ng. 2013.
\newblock \href
  {https://proceedings.neurips.cc/paper/2013/hash/b337e84de8752b27eda3a12363109e80-Abstract.html}
  {Reasoning with neural tensor networks for knowledge base completion}.
\newblock In \emph{Advances in Neural Information Processing Systems 26: 27th
  Annual Conference on Neural Information Processing Systems 2013. Proceedings
  of a meeting held December 5-8, 2013, Lake Tahoe, Nevada, United States},
  pages 926--934.

\bibitem[{Speer et~al.(2017)Speer, Chin, and Havasi}]{DBLP:conf/aaai/SpeerCH17}
Robyn Speer, Joshua Chin, and Catherine Havasi. 2017.
\newblock \href {http://aaai.org/ocs/index.php/AAAI/AAAI17/paper/view/14972}
  {Conceptnet 5.5: An open multilingual graph of general knowledge}.
\newblock In \emph{Proceedings of the Thirty-First {AAAI} Conference on
  Artificial Intelligence, February 4-9, 2017, San Francisco, California,
  {USA}}, pages 4444--4451. {AAAI} Press.

\bibitem[{Wang et~al.(2021)Wang, Wang, Huang, You, Leskovec, and
  Kuo}]{DBLP:conf/ijcnn/WangWHYLK21}
Bin Wang, Guangtao Wang, Jing Huang, Jiaxuan You, Jure Leskovec, and
  C.{-}C.~Jay Kuo. 2021.
\newblock \href {https://doi.org/10.1109/IJCNN52387.2021.9534355} {Inductive
  learning on commonsense knowledge graph completion}.
\newblock In \emph{International Joint Conference on Neural Networks, {IJCNN}
  2021, Shenzhen, China, July 18-22, 2021}, pages 1--8. {IEEE}.

\bibitem[{Wu et~al.(2022)Wu, Li, Zhang, and Wu}]{DBLP:conf/acl/WuLZW22}
Sixing Wu, Ying Li, Dawei Zhang, and Zhonghai Wu. 2022.
\newblock \href {https://doi.org/10.18653/v1/2022.findings-acl.30} {{KSAM:}
  infusing multi-source knowledge into dialogue generation via knowledge source
  aware multi-head decoding}.
\newblock In \emph{Findings of the Association for Computational Linguistics:
  {ACL} 2022, Dublin, Ireland, May 22-27, 2022}, pages 353--363. Association
  for Computational Linguistics.

\bibitem[{Yao et~al.(2019)Yao, Mao, and
  Luo}]{DBLP:journals/corr/abs-1909-03193}
Liang Yao, Chengsheng Mao, and Yuan Luo. 2019.
\newblock \href {http://arxiv.org/abs/1909.03193} {{KG-BERT:} {BERT} for
  knowledge graph completion}.
\newblock \emph{CoRR}, abs/1909.03193.

\bibitem[{Yu et~al.(2022)Yu, Zhu, Qin, Zhang, Zhao, and
  Jiang}]{DBLP:conf/acl/00020QZ0022}
Wenhao Yu, Chenguang Zhu, Lianhui Qin, Zhihan Zhang, Tong Zhao, and Meng Jiang.
  2022.
\newblock \href {https://doi.org/10.18653/v1/2022.findings-acl.149}
  {Diversifying content generation for commonsense reasoning with mixture of
  knowledge graph experts}.
\newblock In \emph{Findings of the Association for Computational Linguistics:
  {ACL} 2022, Dublin, Ireland, May 22-27, 2022}, pages 1896--1906. Association
  for Computational Linguistics.

\bibitem[{Zhang et~al.(2020{\natexlab{a}})Zhang, Khashabi, Song, and
  Roth}]{DBLP:conf/ijcai/ZhangKSR20}
Hongming Zhang, Daniel Khashabi, Yangqiu Song, and Dan Roth.
  2020{\natexlab{a}}.
\newblock \href {https://doi.org/10.24963/ijcai.2020/554} {Transomcs: From
  linguistic graphs to commonsense knowledge}.
\newblock In \emph{Proceedings of the Twenty-Ninth International Joint
  Conference on Artificial Intelligence, {IJCAI} 2020}, pages 4004--4010.
  ijcai.org.

\bibitem[{Zhang et~al.(2020{\natexlab{b}})Zhang, Liu, Pan, Song, and
  Leung}]{DBLP:conf/www/ZhangLPSL20}
Hongming Zhang, Xin Liu, Haojie Pan, Yangqiu Song, and Cane~Wing{-}Ki Leung.
  2020{\natexlab{b}}.
\newblock \href {https://doi.org/10.1145/3366423.3380107} {{ASER:} {A}
  large-scale eventuality knowledge graph}.
\newblock In \emph{{WWW} '20: The Web Conference 2020, Taipei, Taiwan, April
  20-24, 2020}, pages 201--211. {ACM} / {IW3C2}.

\bibitem[{Zhang et~al.(2021)Zhang, Liang, Jatowt, Lei, Wei, Jiang, and
  Yang}]{DBLP:conf/emnlp/ZhangLJLWJY21}
Yao Zhang, Hongru Liang, Adam Jatowt, Wenqiang Lei, Xin Wei, Ning Jiang, and
  Zhenglu Yang. 2021.
\newblock \href {https://doi.org/10.18653/v1/2021.emnlp-main.276} {{GMH:} {A}
  general multi-hop reasoning model for {KG} completion}.
\newblock In \emph{Proceedings of the 2021 Conference on Empirical Methods in
  Natural Language Processing, {EMNLP} 2021, Virtual Event / Punta Cana,
  Dominican Republic, 7-11 November, 2021}, pages 3437--3446. Association for
  Computational Linguistics.

\end{thebibliography}
\bibliographystyle{acl_natbib}

\appendix

\section{Algorithm for Normalizing Tail Events}
\label{app:node_homogenization}

Algorithm~\ref{alg:A} presents the pseudo-code of Normalizing Tail Events in Section~\ref{sec:converting}.

\begin{algorithm}[ht]
% \resizebox{1\columnwidth}{!}{
\caption{Normalizing Tail Events} 
\label{alg:A}  
\hspace*{0.02in}{\bf Input:}
A set of annotations $A$ and relations $R$\\
\hspace*{0.02in}{\bf Output:}
A set of sentences in present tense $FA$
\begin{algorithmic}[1]
\STATE Remove annotations with underscores or none, and get a series of filtered annotations $FA$
\FOR{each $fa \in FA$, $r \in R$}
    \STATE Obtain the dependency tree $dep$ and POS tagging result $pos$ of $fa$ 
    \STATE Find $sub$ node with POS $prp$ and edge $subj$ connected directly to it
    \IF{Position of $sub$ is at the start of $fa$}
        \STATE Remove $sub$ in $fa$
    \ENDIF
    \STATE Find node $verb$ with POS $vb$ in $fa$
    \IF{$r$ $\in$ $[xIntent, xWant, xNeed, oWant]$ AND the first word of $fa$ is \emph{to}}
        \STATE Remove the first \emph{to} of $fa$ 
    \ENDIF
    \STATE Transform node $verb$ in $fa$ to its root form
    \STATE Append $suf$ $\in$ $[-s, -es, -ies, ...]$ to $verb$ based on English grammar
    \IF{$r \in [xAttr, xReact]$}
        \STATE Insert \emph{PersonX is} to the start of $fa$
    \ELSIF{$r$ is $oReact$}
        \STATE Insert \emph{PersonY is} to the start of $fa$
    \ELSIF{$r \in [oWant, oEffect]$}
        \STATE Insert \emph{PersonY} to the start of $fa$
    \ELSE
        \STATE Insert \emph{PersonX} to the start of $fa$
    \ENDIF
\ENDFOR
\STATE Return $FA$
\end{algorithmic}
\end{algorithm}

\section{Implementation Details}

\paragraph{Rel-CSKGC}
We use RoBERTa-large containing 335M parameters as the base model.
We use a maximum sequence length of 100 and batch size of 128.
The Adam optimizer is used for optimization with a learning rate of 2e-5 for RoBERTa-large and a learning rate of 1e-4 for MLP layers.
The warmup proportion is set to 0.1.
We train Rel-CSKGC with 1 NVIDIA RTX 3090 Graphical Card for 5 epochs, and it takes 20 hours to finish the training.

% \vspace{0.4em}

\paragraph{\cometTT{}}
To train \cometTT{}, we use the implentations provided here.~\footnote{\url{https://github.com/atcbosselut/comet-commonsense}}
We use the learning rate of 1.625e-5 and the default values for other parameters.

% \vspace{0.4em}

\paragraph{Generative Commonsense Reasoning}
BART-base is employed as the base model, which contains 140M parameters.
We use a batch size of 128 and use the default values for other parameters.

\end{document}